%% file: ms.tex
\documentclass[final]{cvpr}

\usepackage{times}
\usepackage{epsfig}
\usepackage{graphicx}
\usepackage{amsmath}
\usepackage{amssymb}
\usepackage{booktabs}
\usepackage{caption}
\usepackage{subcaption}

\usepackage[table]{xcolor}

\usepackage[pagebackref=true,breaklinks=true,colorlinks,bookmarks=false]{hyperref}

\usepackage[capitalize]{cleveref}
\crefname{section}{Sec.}{Secs.}
\Crefname{section}{Section}{Sections}
\Crefname{table}{Table}{Tables}
\crefname{table}{Tab.}{Tabs.}

\definecolor{red}{rgb}{1, 0.0, 0.0}

\definecolor{DarkBlue}{rgb}{0.0, 0.0, 0.8}

\newcommand{\extention}[1]{{\color{black} {#1}}}

\def\cvprPaperID{0009} 
\def\confYear{CVARVR 2022}

\begin{document}

\title{SparseFormer: Attention-based Depth Completion Network}  
\author{Frederik Warburg$^1$\\
Technical University of Denmark\\
{\tt\small frwa@dtu.dk}
\and
Michael Ramamonjisoa$^1$\\
Ecole des Ponts\\
{\tt\small michael.ramamonjisoa@enpc.fr}
\and
Manuel López-Antequera\\
Meta\\
{\tt\small mlop@fb.com}
}

\maketitle
\let\thefootnote\relax\footnotetext{$^1$ Work was done during internship at Meta.}
\thispagestyle{empty}

\input{Sections/01abstract.tex}
\input{Sections/02intro.tex}
\input{Sections/03relatedwork.tex}

\input{Sections/04method.tex}

\input{Sections/05experiments.tex}

\input{Sections/06conclusion.tex}

\newpage
{\small
\bibliographystyle{ieee_fullname}
\bibliography{ms}
}

\end{document}

%% file: Sections/01abstract.tex
\begin{abstract}
    Most pipelines for Augmented and Virtual Reality estimate the ego-motion of the camera by creating a map of sparse 3D landmarks. In this paper, we tackle the problem of depth completion, that is, densifying this sparse 3D map using RGB images as guidance.
    This remains a challenging problem due to the low density, non-uniform and outlier-prone 3D landmarks produced by SfM and SLAM pipelines.
    We introduce a transformer block, SparseFormer, that fuses 3D landmarks with deep visual features to produce dense depth. The SparseFormer has a global receptive field, making the module especially effective for depth completion with low-density and non-uniform landmarks.
    To address the issue of depth outliers among the 3D landmarks, we introduce a trainable refinement module that filters outliers through attention between the sparse landmarks.

    

\end{abstract}

%% file: Sections/02intro.tex

\input{Sections/02intro_short}

\extention{The core of the SparseFormer is an attention volume between 3D landmarks projected onto the image plane and visual features extracted from a convolutional decoder. This attention volume describes the similarity between each region in an image and each 3D landmark.
This can be interpreted as an affinity matrix that can be used to interpolate depth to the entire scene in a single step. The global nature of this affinity allows the model to generalize well to different landmark distributions and sparsity levels. Since the number of SfM landmarks is usually low ( $\approx 300$ per frame), the attention mechanism effectively scales linearly with the feature map resolution, allowing us to compute the attention volume without large memory requirements. To accommodate potential outliers in the input depth, we introduce a refinement module that prevents the SparseFormer from propagating noisy information to the interpolation process.
This is achieved by concatenating the depth to the features at those SfM landmarks, and by allowing these depth-aware features to exchange information via a standard transformer.}


\extention{To validate our approach, we first perform an ablation study on the number of 3D points on an indoor dataset. We find that the SparseFormer especially improves performance even for low densities of points. Then we show on a large outdoor dataset that our approach is effective for \textit{very sparse depth completion} with 3D points obtained from a standard SfM pipeline. } 

%% file: Sections/02intro_short.tex
\section{Introduction}


In depth completion, an image is used to guide the densification of a sparse depth map obtained from a sensor (LiDAR, RealSense, etc.) or 3D landmarks from \textit{e.g.} Structure from Motion (SfM). 
Depth completion methods can be divided into (1) \textit{very sparse depth completion} that densify depth from SfM landmarks, (2) \textit{sparse depth completion} that densify depth from LiDAR, and (3) \textit{semi-dense depth completion} that densify depth from \textit{e.g.} RealSense/Kinect. The differentiation is important as SfM landmarks are less uniformly distributed, more error prone and much sparser (typically $ < 0.1 \%$ of pixels have depth) than depth from LiDAR (depth density $\approx 10 \%$) or from depth sensors such as RealSense or Kinect ( depth density $60-90 \%$). Due to the popularity of the KITTI Depth Completion benchmark~\cite{SparseConv_Uhrig173DV} most depth completion methods have focused on \textit{sparse depth completion} from LiDAR data where depth measurements are spatially uniformly distributed. These methods often fail to produce accurate depth for sparser or less evenly distributed data. In this paper, we present the SparseFormer, which is specifically designed for \textit{very sparse depth completion} from SfM landmarks. 

%% file: Sections/03relatedwork.tex
\section{Related work}

Depth completion has shown superior performance compared to monocular depth prediction and higher versatility and efficiency compared to stereo approaches. Despite impressive performance, depth completion is still a challenging task, where the main challenges stem from fusing different modalities (RGB and depth) and handling sparse data. Existing depth completion methods can be divided into three non-mutually exclusive categories: Early fusion, late fusion, and iterative approaches.

\extention{\textbf{Early Fusion} methods fuse image and depth early in the network. The motivation is that the network can dedicate all of its capacity to optimally combine the modalities through training. Early works on depth completion focused on early fusion and tried to make convolutions handle sparse data better. \cite{SparseConv_Uhrig173DV} proposed sparsity invariant convolutions, which normalize the convolutional operator by the number of sparse points in its receptive field. \cite{NConv_Eldesokey18PAMI} proposed to use normalized convolutions to propagate a binary~\cite{NConv_Eldesokey18PAMI, NConv_Eldesokey20BMVC} or continuous learned~\cite{NConv_Eldesokey20CVPR} confidence through the network. These methods strive to make the convolutions handle sparse data better, but do not deal with the computational overhead of convolving regions without any information, which is the case for most regions in \textit{very sparse depth completion}. Other approaches have tried to densify the sparse input depth before feeding it to the network. \cite{Chen18ECCV} proposed to achieve this with nearest neighbor upsampling, while also modelling the distance to the closest sparse point. \cite{VOICED_Wong19ICRA} used scaffolding to densify the sparse input depth. For \textit{sparse and semi-dense depth completion}, where the input depth is rather uniformly distributed, good performance can also be achieved by simple concatenation of image and depth \cite{Sparse2Dense_Ma18ICRA, PeNET_Hu21ICRA, ACDC_Warburg21preprint}. ACDCNet~\cite{ACDC_Warburg21preprint} obtains impressive results by using a channel exchanging network, which allows the network exchange information between different modalities throughout the network, fusing depth from SfM landmarks, depth from a RealSense sensor and an image.}

The downside of early fusion methods is that they do not preserve depth and that they do not perform well when the sparse data is not uniformly distributed or too sparse, which is the focus of this paper. Our main insight is that transformers are more suited for very sparse data than convolutions.

\textbf{Late Fusion} methods explicitly guide the depth completion process by first extracting features from the image and using these features to guide a depth interpolation process. \extention{GuideNet~\cite{GuideNet_Tang19} proposed to imitate guided filtering by learning a spatial variant kernel, ENet~\cite{PeNET_Hu21ICRA} simplified this by an element-wise multiplication of features from input depth and features from an image. RigNet~\cite{RigNet_Yan21Preprint} proposes a repetitive image guided network with an adaptive guidance of depth.} These late fusion approaches have state-of-the-art performance on the KITTI depth completion benchmark~\cite{SparseConv_Uhrig173DV}. We follow this strategy and insert our proposed SparseFormer module in the decoder.

\textbf{Iterative Fusion} methods merge sparse depth and a predicted depth map in depth space. A network predicts initial depth and affinities and the initial depth is updated iteratively using the learned affinities. This merging constrains the network to preserve depth when it is available, while having smoothness between predicted depth and sparse depth. \extention{SPN~\cite{SPN_Lie17NIPS} was the first to propose to predict affinities followed by an iterative diffusion. CSPN~\cite{CSPN_Cheng18CVPR} proposed to predict affinities to the $8$ nearest neighbors in a grid, and showed that the iterative updates can efficiently be computed with convolutions. CSPN++~\cite{CSPN++_Cheng19AAAI} proposed a more efficient diffusion process by using kernels with different strides and predicting the number of iterative update steps. PeNET~\cite{PeNET_Hu21ICRA} proposes to accelerate CPSN++ by better parallelization.
NLSPN~\cite{NLSPN_Park20ECCV} uses deformable convolutions to avoid the grid constraint of previous works. GDC~\cite{GDC_You19ICLR} does not predict affinities, but estimates them in 3D based on local similarity and updates depth in 3D by optimizing a constraint graph. \cite{SparsetoDenseDC_Xiong20ECCV} also models the diffusion in 3D, but uses a graph-convolutional network to update depth of point neighborhoods.}
The downside of these approaches is the iterative update, which is slow, and therefore often constrained to $10-30$ update steps.
The affinities model depth similarity between neighboring points (in 2D or 3D), not allowing information to diffuse globally, \textit{e.g.} observing a sparse point on a plane fronto-parallel to the camera sensor should describe the depth of the entire plane, however, these methods cannot diffuse depth to distant regions as the number of update steps has to be kept rather small for computational reasons. In contrast, the SparseFormer is a global approach that enables the diffusion of depth to the entire scene in one update. 

\textbf{Transformers:} Recent work has shown that transformers outperform convolutional approaches across many tasks such classification~\cite{ViT_Dosovitskiy20ICLR}, object detection~\cite{DETR_Carion20ECCV} and depth prediction~\cite{DPT_21ICCV}. The core of transformers is the attention mechanism that gives the networks a global receptive field. Standard attention scales quadratically with the image size. Multiple recent works have tried to come up with general solutions that can alleviate the huge memory requirement for naive attention. \extention{\cite{ViT_Dosovitskiy20ICLR} proposed to divide the image into patches and perform attention between the patches. Other approaches~\cite{Linformer_Wang20NIPS, reformer_kitaev20iclr, Choromanski20ICLR} propose to approximate the attention volume.}
The nature of the problem of very sparse depth completion allows us to sidestep the issue:
%
Since the number of sparse features with depth, in our setting, is very low, our attention module effectively scales linearly rather than quadratically with the image size. 

%% file: Sections/04method.tex
\begin{figure*}[!htp]
    \centering
    \includegraphics[width=1\textwidth]{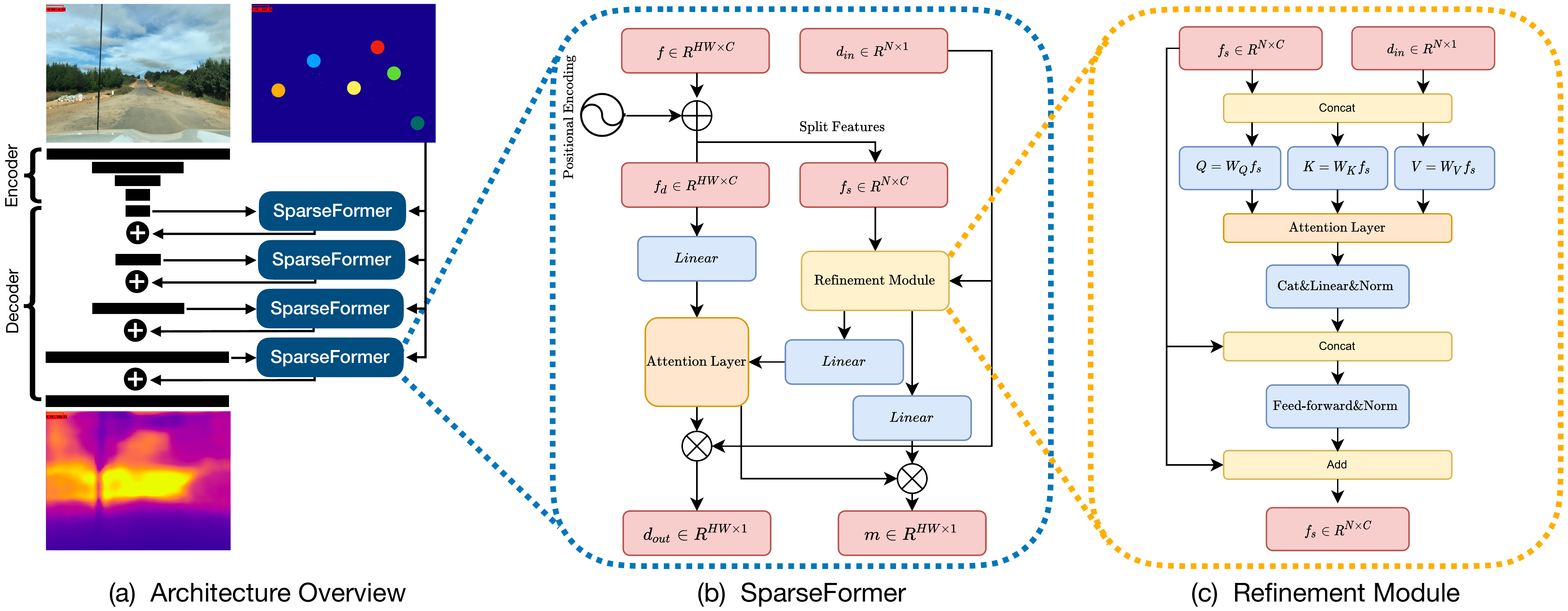}
    \caption{\textbf{(a)~Architecture overview.} We insert $4$ SparseFormers into a depth prediction encoder-decoder architecture~\cite{monodepth2_Godard19ICCV}. \textbf{(b)~SparseFormer.} The SparseFormer takes a convolutional feature map $f$, and $N$ 3D sparse points as input and outputs an interpolated depth map $d_{out}$ and a confidence map $m$. \textbf{(c)~Refinement~module.} The refinement module fuses deep features of SfM points and their associated depth, allowing the features to communicate via attention.}
    \label{fig:model_overview}
\end{figure*}

\section{Method}

The proposed method consists of $T$ SparseFormers that can be inserted in any decoder to enrich the features with global depth information. Each SparseFormer can be viewed as a learned, global diffusion process. 
\cref{fig:model_overview} serves as an overview of our method. 
The SparseFormer takes deep features and 3D landmarks projected into the image plane as input and outputs a dense depth and confidence map, which are used to enrich the convolutional features. 

\paragraph{SparseFormer.} The goal of each SparseFormer is to diffuse depth and confidence from a few 3D points to enrich deep convolutional features. \textit{E.g.} if a sparse point is observed on a plane, we aim to diffuse the known depth to the entire plane. In contrast to iterative methods \cite{SPN_Lie17NIPS,CSPN_Cheng18CVPR,CSPN++_Cheng19AAAI,NLSPN_Park20ECCV}, the SparseFormer can learn to diffuse information to the entire scene - making the module highly flexible and adaptive to different point densities and distributions. 

The SparseFormer takes deep features $f \in R^{H\times W \times C}$ from a traditional convolutional decoder and $N$ 3D points as input, and outputs a dense depth map $d_{out} \in R^{H\times W}$ and a confidence map $m \in R^{H\times W}$. The SparseFormer is specifically designed for situations where $N << HW$ and where the $N$ points can contain outliers. This is in contrast to previous works~\cite{GDC_You19ICLR, NLSPN_Park20ECCV, CSPN_Cheng18CVPR} that assume that the provided sparse depth is always reliable.

\cref{fig:model_overview} (b) gives an overview of a SparseFormer. A positional encoding is concatenated with the dense input features, which are then flattened $f_d \in R^{HW \times C}$. From these features, sparse features $f_s \in R^{N \times C}$ are extracted for each 3D point. The sparse features are refined in order to remove outliers (explained in \cref{sec:refinement} and \cref{fig:model_overview} (c)). We then learn an attention volume $A \in R^{HW \times N}$ that describes the similarity between each feature and the $N$ 3D points.
\begin{equation}
    A = \text{Softmax}\left(\frac{(W_qf_s)(W_kf_d)^T}{\sqrt{C}}\right)
\end{equation}
where $W_q, W_k$ are learnable weights. Since $N$ is small ($< 500$), this attention volume scales approximately linearly with the feature resolution, thus can be computed in high resolution with a relatively low memory footprint.

The attention volume can be interpreted as learned interpolation weights based on feature similarity and positional similarity. We use $A$ to interpolate dense depth $d_{out} \in R^{HW\times1}$ and confidence $m \in R^{HW\times1}$
\begin{align}
    d_{out} &= A d_{in}  \\
    m &= W_{o} (A (W_v f_s))
\end{align}
where $W_{o}, W_v$ are learnable weights. 

The interpolated depth $d_{out}$ and confidence $m$ is merged with the features from the convolutional decoder with a $1\times1$ convolutional layer.

\paragraph{Refinement Module.}\label{sec:refinement} The refinement module has three objectives: (1) add depth information to the deep features, and allow the features to communicate to (2) filter outliers and (3) improve the depth. \cref{fig:model_overview} (c) gives an overview of the refinement module. \extention{First, deep features for each of the 3D points are concatenated with their associated depth to add depth information. Then, these concatenated features are fed through a standard transformer with self-attention. The transformer improves the feature of each 3D point by allowing it to share information between the other landmarks. The transformer can also filter 3D points with erroneous depth, which are common in SfM pipelines due to \textit{e.g.} wrong 2d-2d matches. It can achieve this by learning a mapping that sends these features far away from the rest of the features in the embedding space. This is important in order to not propagate erroneous depth to the interpolation stage.
}

\begin{table*}[]
    \centering
    \begin{tabular}{lll|lllll}
    \toprule
         Architecture &	Authors & \#3D points &REL & RMSE &a1 & a2 & a3 \\
    \midrule
         NLSPN~\cite{NLSPN_Park20ECCV}          &  Park’20  &  2 &  0.300  & 1.152  & 0.393  & 0.697  &  0.879 \\
         SparseFormer   &           &  2 &  \bf{0.161} & \bf{0.626}  & \bf{0.740}  & \bf{0.937}  &  \bf{0.984} \\
    \hline
         NLSPN~\cite{NLSPN_Park20ECCV}          &  Park’20  &  32 &  0.114 & 0.554  & 0.825  & 0.947  &  0.985  \\
         SparseFormer   &           &  32 &  \bf{0.050} & \bf{0.255}  & \bf{0.962}  &  \bf{0.992} & \bf{0.998}   \\
    \hline
         NLSPN~\cite{NLSPN_Park20ECCV}          &  Park’20  &  200 & \bf{0.019}  & \bf{0.136}  &  \bf{0.989} &  \bf{0.998} & 0.999   \\
         SparseFormer   &           &  200 & 0.022  & 0.142  & \bf{0.989}  &  \bf{0.998} & \bf{1.000}   \\
    \hline
         NLSPN~\cite{NLSPN_Park20ECCV}          &  Park’20  &  500 & \bf{0.012}  & \bf{0.092}  & \bf{0.996}  & \bf{0.999}  & \bf{1.000}   \\
         SparseFormer   &           &  500 & 0.014  & 0.104  & 0.994  & \bf{0.999}  & \bf{1.000}   \\
    \bottomrule
    \end{tabular}
    \vspace{0.1cm}
    \caption{\textbf{Results on NYUDv2.} The SparseFormer outperforms NLSPN when few points are available. This is due to the global receptive field of the SparseFormer that diffuses depth to all regions in the image. Both methods use a Resnet34 encoder.}
    \label{tab:nyudv2results}
\end{table*}

\begin{figure*}
    \centering
    \includegraphics[width=\textwidth]{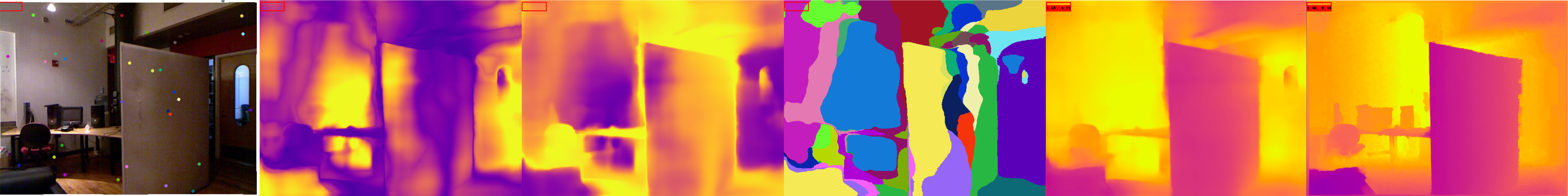}
    \vspace{0.05cm}
    \includegraphics[width=\textwidth]{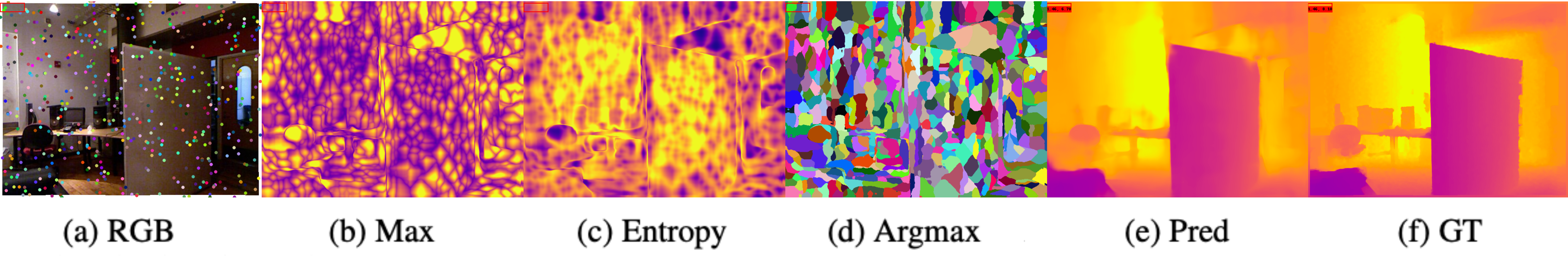}
    \caption{\textbf{Visualization of global depth diffusion.} \textit{From left to right:} RGB image with sparsely sampled ground-truth points emulating 3D landmarks, maximum over attention values, entropy over attention volume, argmax over attention volume color-coded to match to the sparse point in the RGB image for respectively $32$ and $500$ sampled points. The SparseFormer adapts to very different input depth densities. The points are interpolated in a local area thanks to the positional encoding. The interpolation clearly respects boundaries at depth discontinuities. Notice how the interpolation is vertical on walls whereas on fronto-parallel surfaces the interpolation is omni-directional.}
    \label{fig:argmax_visualization}
\end{figure*}

\paragraph{Handling varying levels of depth sparsity.} The number of sparse depth points vary from scene to scene. In order to handle the varying number of point densities and to train in an efficiently batched way, the number of 3D points is fixed. This is done by padding, if the scene contains fewer points than the specified threshold and through random selection if the scene contains more sparse depth points than the specified threshold. 

%% file: Sections/05experiments.tex
\section{Experimental Results}

We insert $4$ SparseFormers in the decoder of the Monodepth2~\cite{monodepth2_Godard19ICCV}. We supervise the final output prediction as well as the interpolated depth map of each SparseFormer, such that our final loss becomes $\sum_s w_s (l_1 + l_2)$ where $s$ iterates over scales and $w_s = (1.0, 0.5, 0.25, 0.15, 0.10)$. We use an Adam optimizer ($\beta_1 = 0.9$, $\beta_2 = 0.999$) and train for $600k$ iterations with learning rate $0.001$, which is decayed by $80 \%$ every $200k$ iterations. We use batch size~$24$.

We train and evaluate our model on two datasets: The NYU Depth dataset and the Mapillary Planet-Scale Depth dataset. For quantitative comparison, \extention{we report the following commonly used metrics~\cite{NLSPN_Park20ECCV}.
\begin{align}
    \text{REL:} &\quad \frac{1}{|\mathcal{V}|} \sum_{v \in \mathcal{V}}\left|\left(d_{v}^{g t}-d_{v}^{p r e d}\right) / d_{v}^{g t}\right| \\
    \text{RSME:} &\quad \sqrt{\frac{1}{|\mathcal{V}|} \sum_{v \in \mathcal{V} \mid}\left|d_{v}^{g t}-d_{v}^{p r e d}\right|^{2}} \\
    \delta_t: &\quad \max \left(\frac{d v_{v}^{g t}}{d_{v}^{p r e d}}, \frac{d_{v}^{p r e d}}{d_{v}^{g t}}\right)<\tau
\end{align}
}

\paragraph{NYU Depth V2.} (NYUDv2)~\cite{SparseConv_Uhrig173DV} is an indoor dataset where ground truth is obtained with a Kinect sensor.

We follow a similar procedure to~\cite{NLSPN_Park20ECCV} and use a subset of $\approx 50k$ images from the official training split and the official test split of $654$ images for evaluation and comparisons. Each image was downsized to a resolution of $320\times240$, and then center-cropped to $304\times228$ pixels.

\cref{fig:argmax_visualization} illustrates the diffusion mechanism for a varying number of available sparse points. Note how the attention mechanism adapts to the number of sparse points. The depth is diffused to large neighborhoods when only few points are available.

\cref{tab:nyudv2results} shows that our model outperforms NLSPN when very few 3D points are available, and matches the predictive performances of NLSPN when more points are available.


\begin{table}[]
    \centering
    \resizebox{\linewidth}{!}{\begin{tabular}{l|l l l l l}
        Architecture  & REL & RMSE & a1 & a2 & a3 \\
        \toprule
        Monodepth2 \cite{monodepth2_Godard19ICCV}         &  0.052 &  15.777 &  0.895 & 0.943  & 0.970\\
        ENet \cite{PeNET_Hu21ICRA}      & 0.019  &  5.469 &  0.975 &  0.988 & 0.992\\
        SparseFormer  & \bf{0.011}  & \bf{4.948}  & \bf{0.989}  & \bf{0.996}  & \bf{0.998}\\
        \bottomrule
    \end{tabular}}
    \caption{\textbf{Results on MPSD.} All methods use a Resnet101 encoder.}
    \label{tab:mpsd_dataset}
\end{table}

\paragraph{MPSD} We first compute SfM landmarks on the Mapillary Planet-Scale Depth (MPSD)~\cite{MPSD_Antequera20ECCV} dataset. The dataset contains $\approx 1$ M images with a large geographical coverage. The SfM landmarks are generated with OpenSfM using SIFT features, thus representing a typical scenario for \textit{very sparse depth completion}. 

\begin{figure}[!htp]
    \centering
    \includegraphics[width=\linewidth]{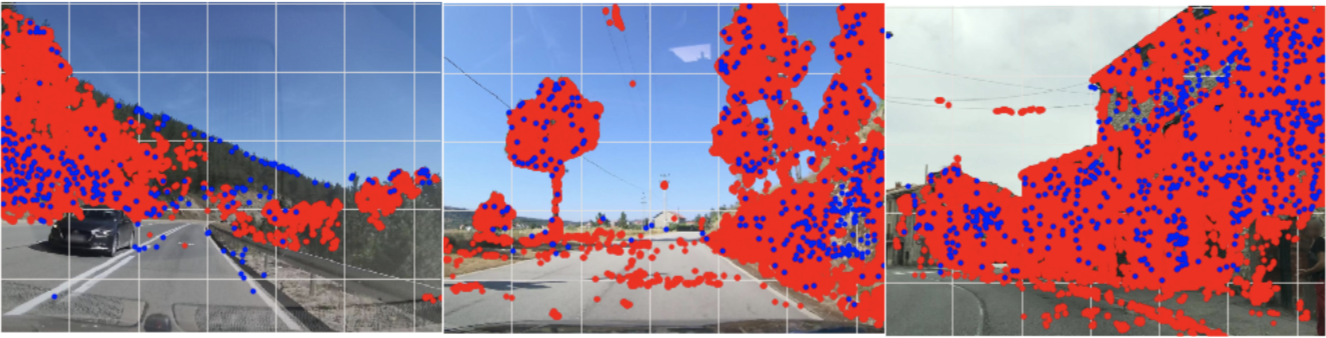}
    \caption{SfM landmarks projected into the camera (blue) and the ground-truth depth (red). }
    \label{fig:mpsdc_sfm_gt}
    \includegraphics[width=\linewidth]{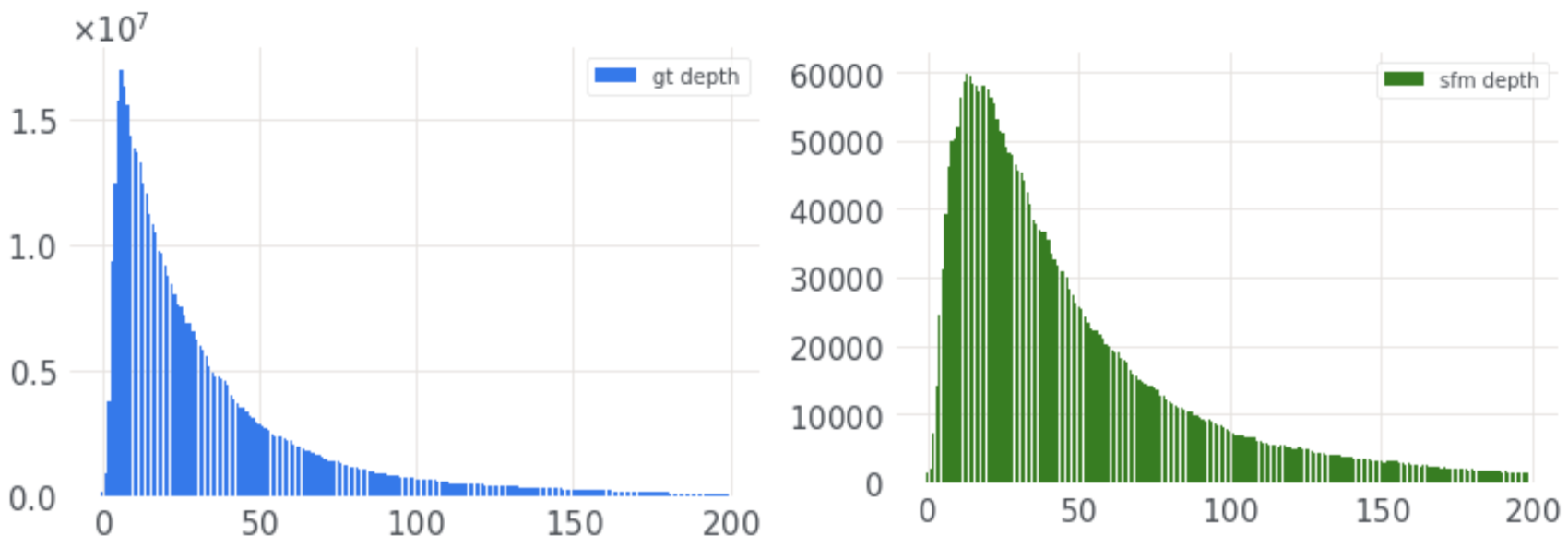}
    \caption{Depth distribution of SfM landmarks and ground-truth in meters. SfM landmarks are on average further away than the ground-truth depth.}
    \label{fig:mpsdc_distribution_sfm_gt}
\end{figure}

\cref{fig:mpsdc_sfm_gt} shows that neither the ground-truth depth nor the input 3D points are uniformly distributed in the image\extention{, but rather  clustered in textured areas and around boundaries.} The average density of the 3D points is $0.1 \%$ ($330$ points per frame), whereas the density of the ground-truth is $16 \%$ ($50k$ points per frame).

\extention{\cref{fig:mpsdc_distribution_sfm_gt} shows the distribution of the SfM depth and the ground-truth depth. The SfM points are on average further away from the camera than the ground-truth depth on this dataset.}

\begin{figure}
     \centering
     \begin{subfigure}[b]{0.32\linewidth}
         \centering
         \includegraphics[width=\linewidth]{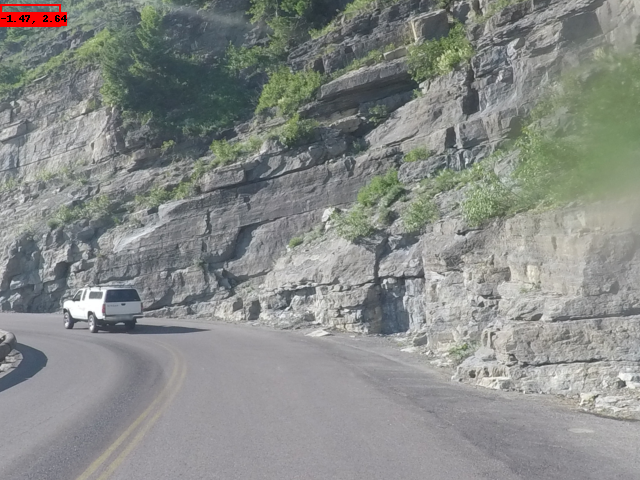}
     \end{subfigure}
     \begin{subfigure}[b]{0.32\linewidth}
         \centering
         \includegraphics[width=\linewidth]{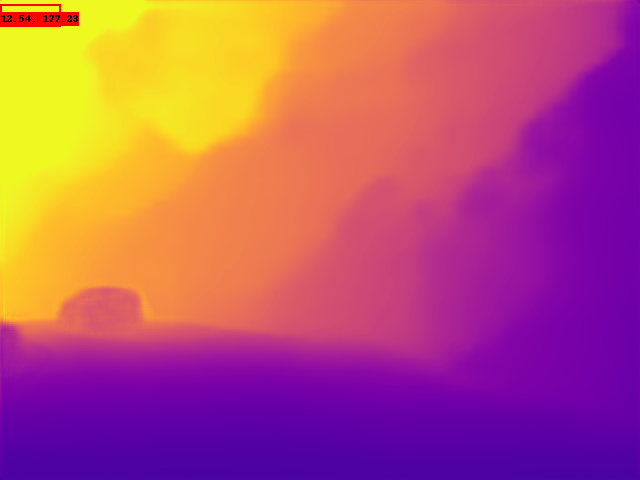}
     \end{subfigure}
     \begin{subfigure}[b]{0.32\linewidth}
         \centering
         \includegraphics[width=\linewidth]{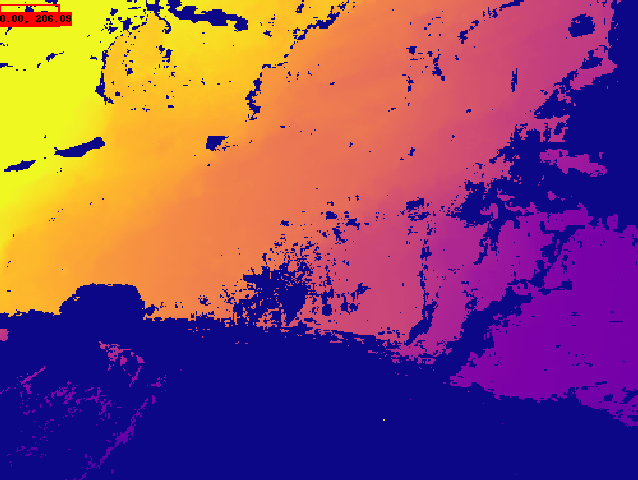}
     \end{subfigure}\\
     
     \begin{subfigure}[b]{0.32\linewidth}
         \centering
         \includegraphics[width=\linewidth]{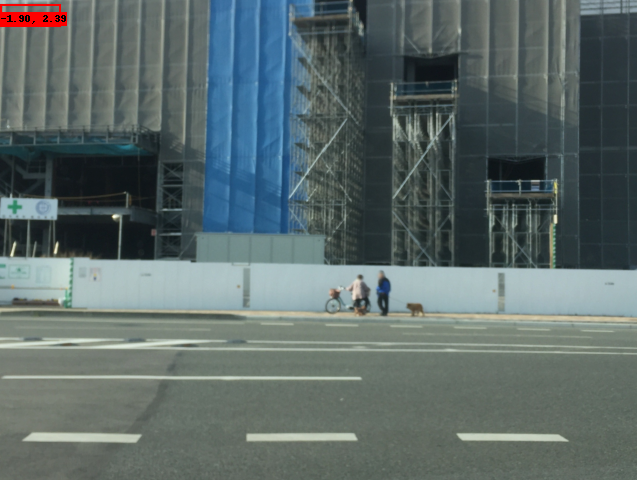}
     \end{subfigure}
     \begin{subfigure}[b]{0.32\linewidth}
         \centering
         \includegraphics[width=\linewidth]{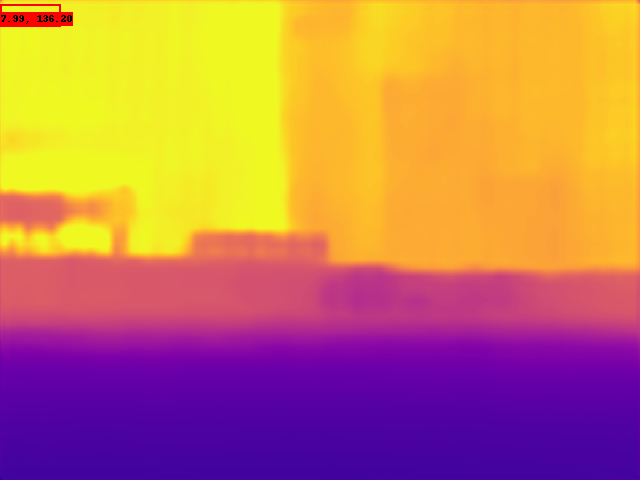}
     \end{subfigure}
     \begin{subfigure}[b]{0.32\linewidth}
         \centering
         \includegraphics[width=\linewidth]{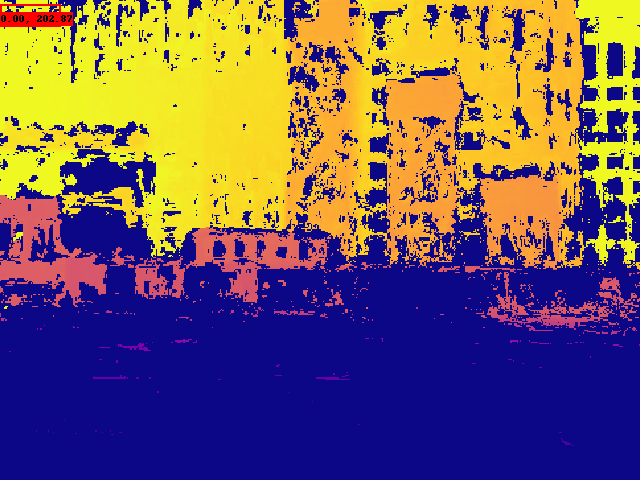}
     \end{subfigure}\\
     
     \begin{subfigure}[b]{0.32\linewidth}
         \centering
         \includegraphics[width=\linewidth]{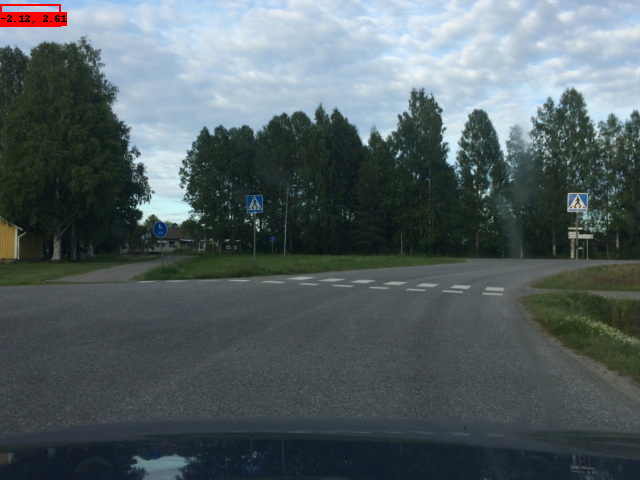}
         \caption{RGB}
     \end{subfigure}
     \begin{subfigure}[b]{0.32\linewidth}
         \centering
         \includegraphics[width=\linewidth]{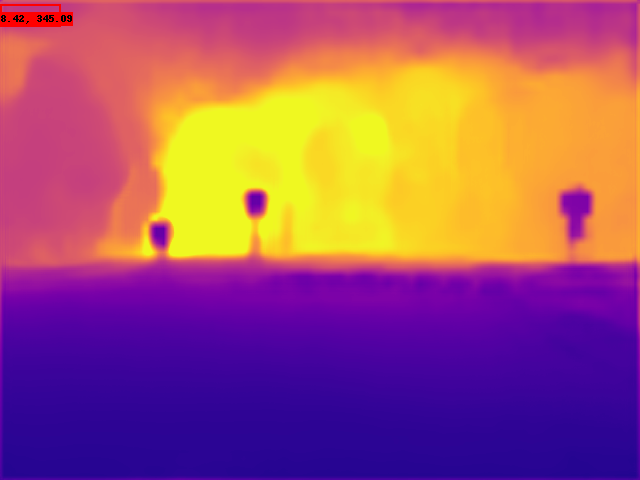}
         \caption{Pred}
     \end{subfigure}
     \begin{subfigure}[b]{0.32\linewidth}
         \centering
         \includegraphics[width=\linewidth]{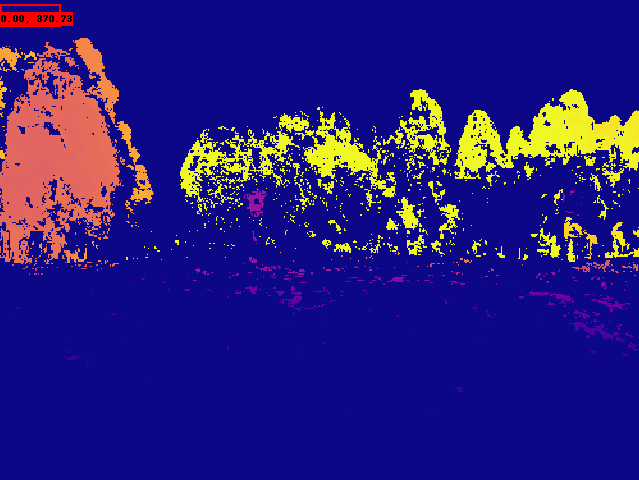}
         \caption{GT}
     \end{subfigure}
        \caption{\textbf{Qualitative results on MPSD.} The SparseFormer learns to produce accurate depth maps using an image and 3D points as input and semi-dense depth as supervision.}
        \label{fig:mpsd_depth_maps}
\end{figure}

We compare against two recent methods, which use a Resnet101 as backbone.
(1) Monodepth2~\cite{monodepth2_Godard19ICCV} which we train supervisedly, and adapt to depth completion by concatenating the RGB and sparse depth inputs.
(2) ENet~\cite{PeNET_Hu21ICRA}, which is one of the top performing methods on the KITTI depth completion dataset.
\cref{tab:mpsd_dataset} shows that the SparseFormer outperforms both baselines due to its global receptive field and the communication between the sparse features.

\cref{fig:mpsd_depth_maps} shows qualitative results from the SparseFormer. The SparseFormer produces accurate depth maps using a few sparse SfM landmarks and RGB images.



%% file: Sections/06conclusion.tex
\section{Conclusion}

We propose a transformer module for \textit{very sparse depth completion}. The SparseFormer is specifically designed for low density, non-uniformly distributed and outlier prone 3D depth. The SparseFormer consists of (i) an interpolation module that globally interpolates 3D points based on visual similarity and position, which allow the network to globally diffuse depth. (ii) A refinement module that fuses depth and visual features and filters outliers through cross-attention of the sparse points. We demonstrate on both indoor and outdoor datasets that the proposed SparseFormer is effective in low density depth completion.